\documentclass[fleqn,10pt]{wlscirep}
\usepackage[utf8]{inputenc}
\usepackage[T1]{fontenc}
\usepackage{graphicx}%
\usepackage{multirow}%
\usepackage{hyperref}
\usepackage{xcolor}%
\title{RSTNet: Enhancing Small-Target Recognition in Noisy SAR Imagery via Robust Feature Learning and Distribution-Aware Regression}

\author[1]{Xiaojing Zhao}
\author[2]{Shiyang Li}
\author[2]{Zenan Chu}
\author[3,4]{Ying Zhang}
\author[2]{Peinan Hao}
\author[5]{Tianzi Yan}
\author[6]{Jiajia Chen}
\author[2,*]{Huicong Ning}

\affil[1]{College of Mathematics and Information Science, Anyang Institute of Technology, Anyang, 455000, China}

\affil[2]{School of Computer Science and Information Engineering, Anyang Institute of Technology, Anyang, 455000, China}

\affil[3]{College of Electronic Information and Optical Engineering, Nankai University, Tianjin, 300350, China}

\affil[4]{Tianjin Key Laboratory of Optoelectronic Sensor and Sensing Network Technology, Tianjin, 300350, China}

\affil[5]{Beijing Xiao Mi Mobile Software Co., Ltd., Beijing, 100080, China}

\affil[6]{Anyang Embedded Intelligent Robot Co., Ltd., Anyang, 455000, China}

\affil[*]{20230005@ayit.edu.cn}

\begin{abstract}
SAR supports all-day-and-night oceanic observation, yet vessel identification from SAR images is hampered by speckle noise, intricate land-sea backgrounds and dim miniature vessels, yielding numerous false identifications and missed targets. We develop an SAR-adaptive stable detection model \textbf{\textit{RSTNet}} based on YOLOv8. A large-kernel channel-separated denoising unit eliminates noise and reserves delicate vessel features; parallel patch-aware attention enhances multi-scale feature extraction for miniature objects; NWD loss substitutes conventional IoU loss to achieve accurate bounding box regression. The proposed model outperforms the original YOLOv8 on the SSDD dataset with 97.0\% precision, 95.1\% recall and 98.9\% mAP@0.5. Validations on the HRSID dataset verify its favorable generalization capacity for coastal miniature vessels. Therefore, our work delivers an effective technical scheme for ocean observation imaging with noisy miniature targets. The source code is available at \url{https://github.com/renhcmhx/SAR.git.}\\

\end{abstract}
\begin{document}

\flushbottom
\maketitle
%
%
\thispagestyle{empty}

\section{Introduction}
Synthetic aperture radar (SAR) leverages coherent imaging and pulse compression to deliver all-day, all-weather observation, and remains effective under darkness and low illumination. Its robustness to lighting conditions, together with rich scattering and polarization cues, makes SAR a practical sensing modality for a wide range of civilian applications, including urban planning, environmental monitoring, disaster assessment, and maritime surveillance \cite{I1,I2}. Among these, ship detection provides critical situational awareness for maritime authorities operating across open oceans, busy waterways, and major ports, where timely and accurate localization is essential for traffic management and safety monitoring \cite{I3,I4}.

\begin{figure*}[t]
    \centering
    \includegraphics[width=0.95\linewidth]{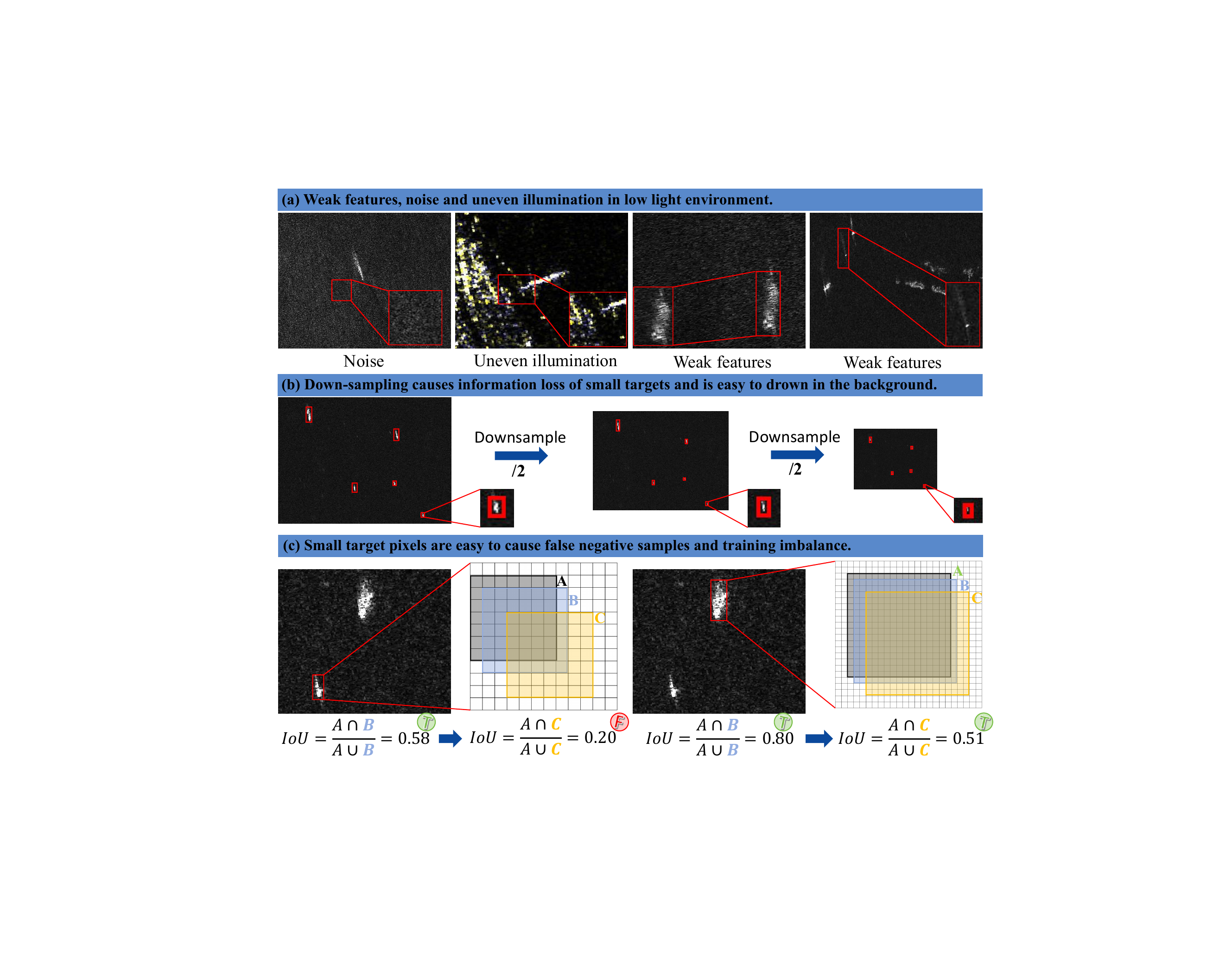}
    \caption{Visualization of motivation. From (a) to (c) represent in turn the 3 challenges of existing methods. (a) Weak features, noise and uneven illumination in low light environment. (b) Down-sampling causes information loss of small targets and is easy to drown in the background. (c) Small target pixels are easy to cause false negative samples and training imbalance.}
    \label{challenge}
\end{figure*}

Early SAR ship detection methods typically rely on intensity contrast between ship returns and surrounding sea clutter. A representative family of approaches models clutter statistics and sets adaptive thresholds to maintain a constant false alarm rate (CFAR), thereby separating ship targets from complex backgrounds \cite{I7}. These methods can perform reliably under relatively calm and stationary sea states, but their accuracy degrades markedly in adverse conditions such as strong winds and heavy rain, where clutter becomes highly non-stationary and exhibits non-Gaussian behavior \cite{I8}. In such cases, mismatched clutter models lead to biased threshold estimation, which in turn increases false alarms and missed detections. Meanwhile, modern SAR systems continue to improve spatial resolution, producing images with richer structural details but also more complex noise patterns and artifacts. As a result, handcrafted feature pipelines tailored to specific scenes often generalize poorly when the acquisition geometry, sea state, or sensor characteristics change.

Deep learning offers an alternative by learning hierarchical representations directly from data, capturing both local textures and high-level semantics within a unified framework \cite{I9,add1,add2,add3}. This data-driven paradigm has repeatedly shown that targeted architectural augmentation and carefully designed objectives can substantially improve robustness under difficult inputs, a principle that has been validated across a range of vision tasks beyond detection, including controllable generation, editing, and long-horizon synthesis \cite{shen2024imagpose,shen2024advancing,shen2025boosting}. Motivated by these advances, CNN-based SAR ship detectors have attracted growing attention \cite{I10,I11}. Existing detectors are commonly grouped into two-stage and one-stage paradigms (see Section~\ref{section:Anchor-based} and Section~\ref{section:Anchor-free}). Two-stage methods often provide strong accuracy but incur higher latency due to proposal generation and refinement, whereas one-stage methods typically enable faster inference with competitive performance, making them attractive for time-sensitive maritime applications \cite{DBW-YOLO}. This trend has driven continuous refinement of one-stage designs, where improvements are increasingly achieved through scenario-aware pre-processing, feature enhancement, and loss re-weighting, which echoes the broader idea of building controllable and reliable learning systems through modular interventions \cite{shen2025imaggarment,shen2025imagdressing}.

Despite steady progress, false positives and missed detections remain prevalent in challenging SAR scenarios, particularly for small ships and cluttered backgrounds. First, many maritime scenes are captured under poor illumination or night-time operating conditions, where target responses are weak, background noise is prominent, and intensity variations are uneven, collectively reducing the discriminability of ship features and complicating detection (Fig.~\ref{challenge}(a)). Second, to control computational cost, most detectors employ repeated down-sampling. However, small ships in SAR imagery often appear with weak signals and blurred contours, so aggressive resolution reduction can discard critical spatial cues. This issue is further amplified by the low contrast and limited physical texture of SAR imagery, which can cause small targets to be submerged in clutter (Fig.~\ref{challenge}(b)). Third, the extreme scale imbalance in SAR images poses a localization and supervision challenge. Ships may occupy only a few pixels, so even minor localization errors can sharply reduce overlap measures and impair positive sample assignment during training. This reduces effective supervision for small objects, exacerbates sample imbalance, and biases optimization toward larger and easier targets (Fig.~\ref{challenge}(c)). These observations are consistent with a recurring pattern in vision systems, where small-object fidelity and stable supervision often require explicit mechanism design rather than relying solely on generic backbones \cite{shen2025imagharmony,shenlong}.

To address these challenges, we build on YOLOv8 \cite{I32} and propose RSTNet, a dedicated SAR ship detection framework that improves robustness to noise, enhances small-object representation, and stabilizes box regression. We choose YOLOv8 as the foundation because its anchor-free design, lightweight feature modules, and dynamic label assignment have shown strong efficiency and deployment friendliness, while still leaving room for scenario-specific adaptation. First, considering the heavy speckle and clutter characteristics of SAR imagery, we introduce a channel-independent denoising module before feature extraction to suppress irrelevant responses and preserve informative structures, which is especially important for small ships. Second, to alleviate the loss of small-object cues under multi-scale down-sampling, we design a PPA-based feature enhancement mechanism that strengthens multi-level feature aggregation and improves attention to small targets. Third, to reduce training instability caused by scale imbalance and imperfect positive sample assignment, we adopt a bounding-box regression loss based on normalized Wasserstein distance, which measures similarity under complex box distributions and improves localization generalization.

The contributions of this paper are summarized as follows.
\begin{itemize}
    \item We propose RSTNet, an improved YOLOv8-based SAR ship detector that jointly addresses SAR noise and small-target detection challenges, achieving strong average precision across diverse scenes.
    \item We integrate two complementary modules, CID and PPA, to reduce cross-channel noise interference and enhance small-scale feature representation through attention-aware multi-level fusion.
    \item We introduce an NWD-based regression loss to improve convergence stability, generalization, and small-object localization by measuring bounding-box similarity via Gaussian distribution modeling.
    \item We conduct extensive experiments on two SAR ship detection benchmarks with different resolutions, and provide quantitative results and visual analyses to validate the effectiveness of the proposed method.
\end{itemize}

\begin{figure*}[t]
    \centering
    \includegraphics[width=0.95\linewidth]{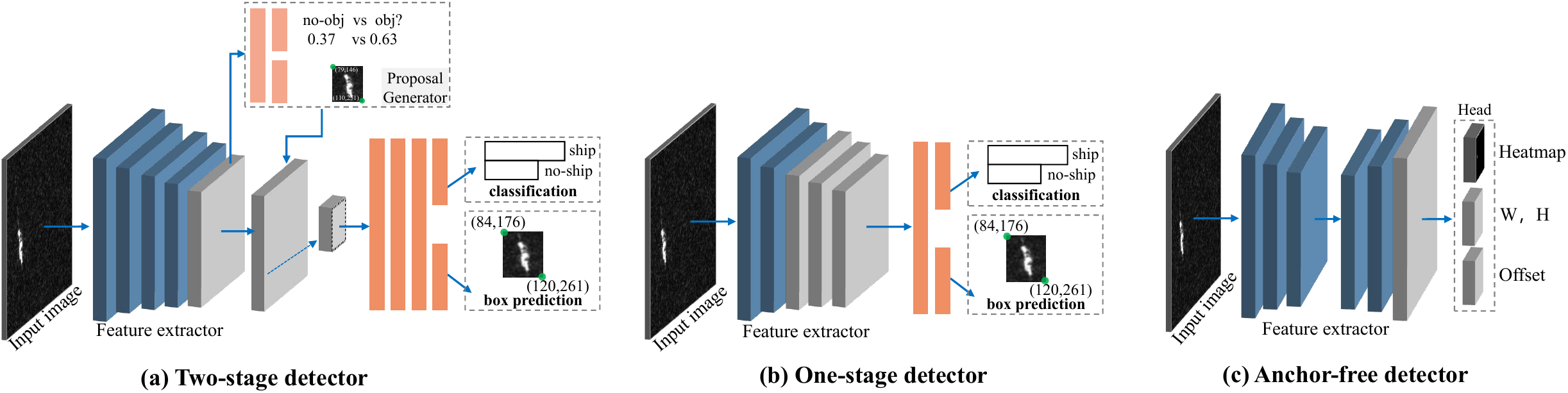}
    \caption{Different types of detectors. From (a) to (c) represent two-stage detector, one-stage detector and anchor-based detector.}
    \label{releated work}
\end{figure*}

\section{Related Work}
With the rapid progress of convolutional neural networks, deep learning has become the dominant paradigm for ship detection in synthetic aperture radar (SAR) imagery, including oriented ship detection. Most existing approaches are developed by adapting general-purpose object detectors to SAR characteristics such as speckle noise, low contrast, and large scale variation. Depending on whether the detector relies on predefined anchors, prior works can be broadly grouped into anchor-based and anchor-free methods.

\subsection{Anchor-based Detectors}
\label{section:Anchor-based}
Anchor-based detectors introduce a set of predefined anchor boxes (or priors) and learn to classify and regress offsets relative to these anchors. They are commonly divided into two-stage and one-stage pipelines. Two-stage detectors (Fig.~\ref{releated work} (a)) typically generate region proposals and then refine them with RoI features. Faster R-CNN \cite{FasterRcnn} is a representative two-stage framework, and subsequent SAR-oriented variants improve performance through stronger backbones, proposal generation, loss design, anchor strategies, and post-processing. Many efforts enhance multi-scale representation with feature pyramids and fusion modules \cite{R198,R199,R200,R42,R62,R63,R69,R91,R110}, or improve feature selectivity via attention mechanisms \cite{R17,R29,R62,R122,R137}. For example, HR-SDNet \cite{R69} adopts a parallel high-resolution feature pyramid to leverage both high- and low-resolution feature streams for SAR ship detection. RC-ARPN \cite{R91} constructs a fine-grained pyramid in a top-down manner with receptive-field segmentation and convolutional attention, improving the representation of high-aspect-ratio ships. DenseFPN \cite{R110} introduces dense connections across pyramid levels and applies differentiated processing to shallow and deep layers to better exploit hierarchical feature diversity.

One-stage anchor-based detectors (Fig.~\ref{releated work} (b)) remove the explicit proposal stage and perform dense prediction over feature maps, making them more suitable for real-time deployment. Representative backbones include YOLO, SSD, and RetinaNet. A large body of work tailors these frameworks to SAR ship detection, mainly by enhancing multi-scale features and improving small-object sensitivity. DWB-YOLO \cite{DBW-YOLO} strengthens feature extraction using deformable convolutions and incorporates attention to better capture coastal and small vessels. CSD-YOLO \cite{CSD-YOLO} introduces an attention-guided feature pyramid and improves scale fusion to reduce feature loss for small ships. SSS-YOLO \cite{R119} redesigns the feature extraction network and uses a bidirectional fusion strategy to enhance both spatial and semantic cues for small targets. Other variants improve SSD via feature fusion and lightweight attention modules \cite{R149}, deconvolution-based pyramid enhancement \cite{R99}, multi-source input fusion \cite{R113}, and prediction head refinement \cite{R114}.

\subsection{Anchor-free Detectors}
\label{section:Anchor-free}
Anchor-free detectors (Fig.~\ref{releated work} (c)) remove predefined anchor boxes and directly predict object locations from feature maps, typically through keypoint estimation or per-pixel regression. FCOS is a representative anchor-free framework that formulates detection as dense prediction, regressing distances from each pixel to the object boundaries while predicting class labels. Building on this idea, AFD \cite{R167} proposes a category-location module that uses guidance vectors from the classification branch to improve localization regression, and redesigns classification and regression targets to alleviate training ambiguity. For SAR imagery, R-FCOS \cite{R180} replaces IoU-based sample assignment with a statistical-property-driven segmentation strategy tailored to SAR characteristics, and further improves feature extraction via single-resolution convolution, multi-resolution fusion, and feature pyramid modules. Overall, anchor-free designs are attractive for SAR ship detection due to simpler label assignment and reduced sensitivity to anchor hyperparameters, and recent studies continue to explore SAR-specific sample selection and multi-scale feature enhancement within this paradigm.

\subsection{Attention mechanism}
Attention mechanisms advance image processing by adaptively focusing on critical features via attention designs. \cite{Visual_Computer_2} proposes an attention-guided network for infrared small target detection, integrating a target attention mechanism (generating attention maps via global pooling and adaptive correction), a multi-resolution feature interaction mechanism, and a contextual feature extraction mechanism, all guided by attention. \cite{DAP-Net} proposes DAP-Net, a dual-channel SAR recognition network incorporating squeeze-and-excitation networks (SENet) and bottleneck attention modules (BAM), with the DCSE module using SENet (a channel attention mechanism) and the improved C3-SCB module integrating BAM (a spatial-channel attention module). \cite{EAPT} proposes EAPT for vision transformers, featuring Deformable Attention (adaptive attention-based patch coverage), an En-DeC module for attention-driven global inter-patch communication, and MCMD for attention-compatible position encoding. \cite{SAT-Net} proposes SAT-Net for fundus image enhancement, integrating window-based self-attention (local spatial attention) and channel self-attention (critical feature emphasis) in its Transformer-based Attention Fusion Module (TAFM). Collectively, these methods adopt attention mechanisms as core components to address image processing challenges.
\section{Methodology}
In this section, we present a comprehensive overview of the proposed method and the implementation details of each structure and key components. The overall architecture consists of four parts: data processing, feature extraction, feature fusion and head prediction. First, we present the overall architecture of the network. Next, we elaborate on the key components of the network in detail. Finally, we introduce the loss function adopted in this paper.
\subsection{Overall architecture of the RSTNet}
\begin{figure*}[t]
    \centering
    \includegraphics[width=0.85\linewidth]{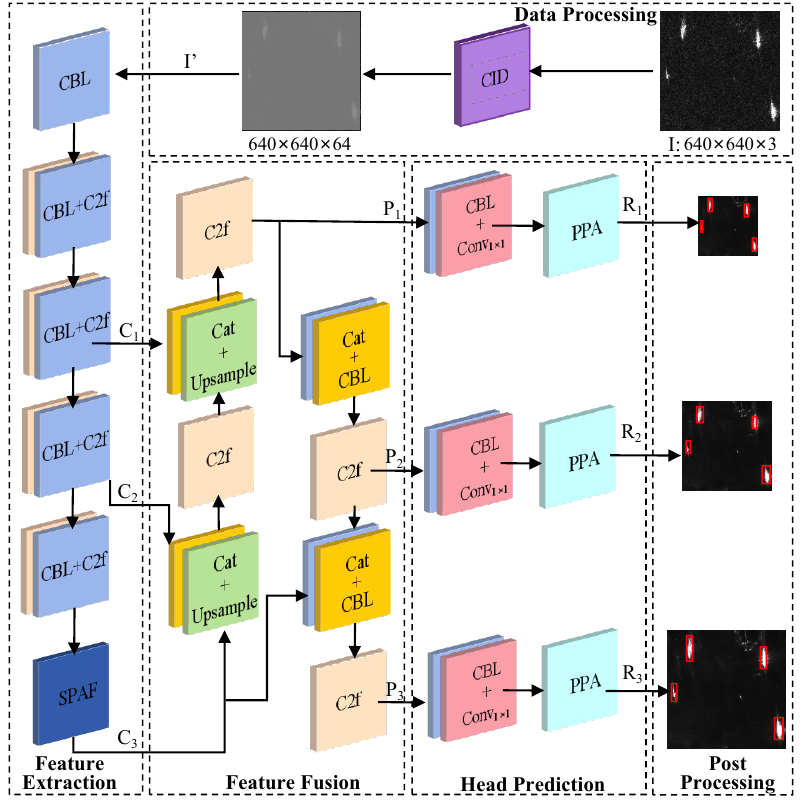}
    \caption{The network architecture of RSTNet. The CBL represents Conv, Batch normalization and SiLU. SPAF denotes SPPF followed by PPA. Among the framework, PPA and CID are the improvements we made.}
    \label{method}
\end{figure*}

To more effectively detect small ships in complex backgrounds, we propose the optimized {RSTNet} for small-scale ships and complex scenes, which can maintain the good performance of small-scale ships, while the algorithm’s primary flow is depicted in Fig.~\ref{method}. Firstly, the geometrically transformed SAR images join the CID and then enter the feature extraction network consisting of CBL blocks, C2f modules and an SPAF block for feature extraction. After that, these features will enter the integration layer for feature fusion, producing better spatial and semantic data, and the feature maps (P1, P2, P3) of various scales will enter the prediction head. Finally, the results are continuously optimized by the NWDLoss function. In the testing phase, the filtered results are additionally required processing via non-maximum suppression (NMS). The variable relationship between the input and output of the proposed {RSTNet} is shown in Tab.~\ref{tab1}.

\begin{table}[]
\caption{Variable relationship between input and output of RSTNet.}
\centering
\begin{tabular}{cccc }
\hline
Task                                                      & Module               & Input    & Output              \\ \hline
Data Processing                                           & CID                  & I        & I’                  \\
Feature Extraction                                        & CBL, C2f, \textbf{SPAF}         & I’       & $C_1, C_2, C_3$            \\
Feature Fusion                                            & Cat, Upsample, C2f, CBL & $C_1, C_2, C_3$ & $P_1, P_2, P_3$            \\
Head Prediction                                           & CBL, Conv1$\times$1, \textbf{PPA}      & $P_1, P_2, P_3$ & $R_1, R_2, R_3$            \\
                                 & Training:            & $R_1, R_2, R_3$ & \textit{$\mathbf{L_{NWD}}$$+L_{BCE}$}  \\
\multirow{-2}{*}{Post Processing} & Inference:           & $R_1, R_2, R_3$ & \textit{B(x,y,w,h)} \\ \hline
\multicolumn{4}{l}{The bold modules are our improvements.} \\ 
\end{tabular}
 \label{tab1}
\end{table}

\subsection{Components}
The {RSTNet} method mainly consists of the following four components, {YOLOv8n}, CID, PPA and NWDLoss as follows:

(1) {YOLOv8n}: {YOLOv8} is one of the YOLO detectors launched by the Ultralytics team in 2023 \cite{M1}. It is well known for its efficient real-time object detection ability, exhibiting fast inference speed and low computational complexity, which makes it ideal for real-time applications and presents excellent performance in most datasets \cite{M27}. According to the different model sizes, it is divided into five models, namely, {YOLOv8n}, {YOLOv8s}, {YOLOv8m}, {YOLOv8l} and {YOLOv8x} \cite{M26}. Compared with the {s/m/l/x} model, {YOLOv8n} has a more compact model size and lower computing requirements, which is very suitable for scenarios with limited computing resources. In addition, {YOLOv8n} tends to outperform its counterparts in terms of processing speed, making it well suited for scenarios with high real-time capability requirements. Therefore, {YOLOv8n} is selected as the foundational model for {RSTNet} method in this paper.

\begin{figure*}[t]
    \centering
    \includegraphics[width=0.55\linewidth]{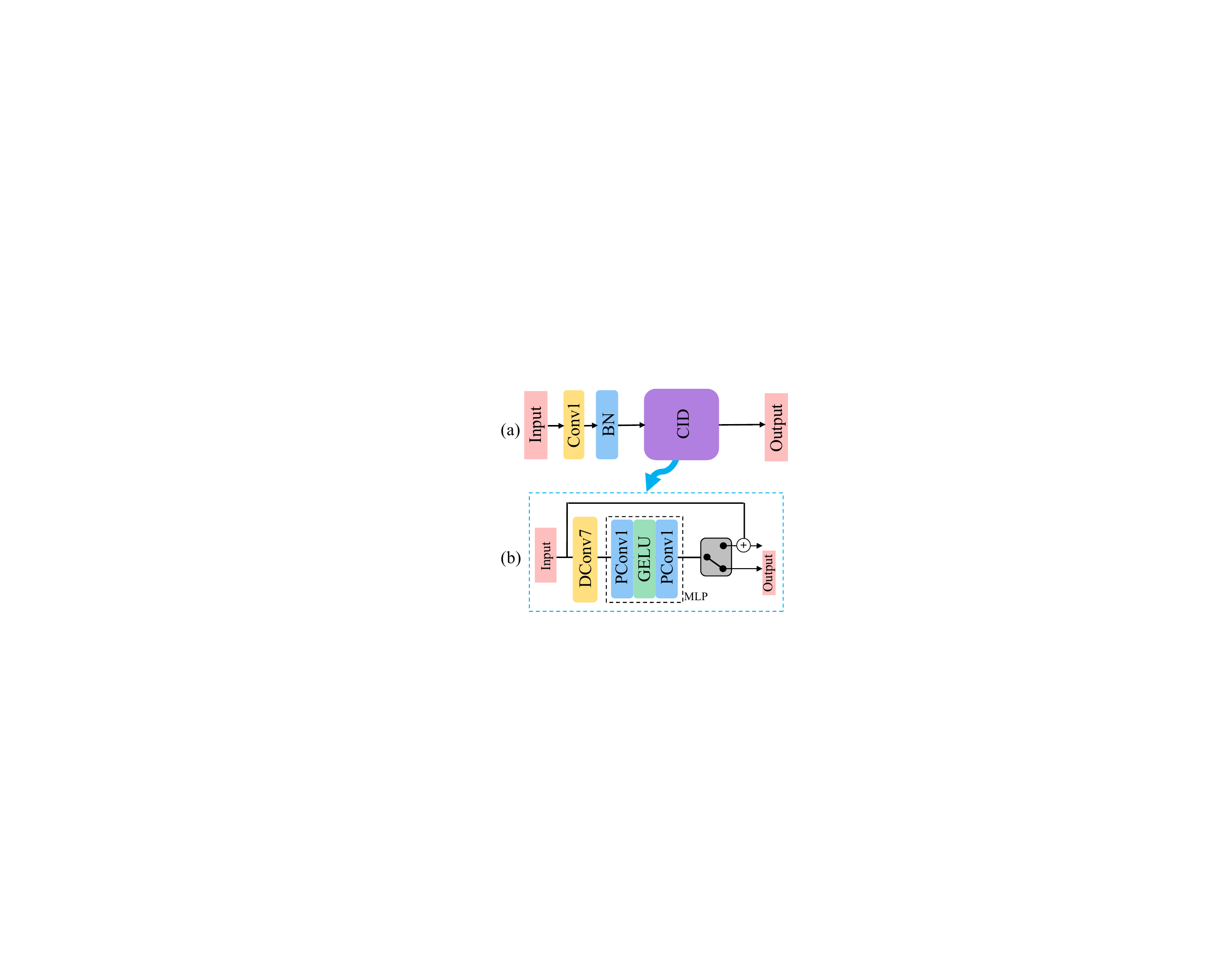}
    \caption{Detailed architectures of introduced task-specific blocks: (a) Channel-Independent Denoising(CID) module. (b) Original CID block.}
    \label{CID}
\end{figure*}

(2) CID: CID was initially proposed by Jin et al. \cite{M2}. Unlike traditional convolution, which employs small kernels, CID introduces a deep convolution with a large kernel and a channel-independent attention mechanism, allowing the network to expand the receptive field in the denoising process and isolate information exchange between different channels. As shown in Fig.~\ref{CID} (b), the original CID module provides two output modes: (1) channel-independent residual output; (2) Channel dependent shortest path.

Due to the interference of complex noise and background, {YOLOv8n} cannot fully solve the problem of small ship recognition. The prior knowledge of SAR image noise is formed based on the following aspects: (1) the noise is not completely random in space, and the noise of adjacent pixels is correlated; (2) sea clutter itself has certain texture and statistical characteristics, which are mixed with noise to form complex interference. Inspired by large kernel convolution with the attention mechanism to enhance important information, we use the CID module that collaborates with large kernel deep convolution and channel attention to enhance the input image. This module forces the network to perform feature optimization from two orthogonal dimensions, spatial and channel. The large kernel convolution is responsible for a wide range of information aggregation and context modeling in the spatial dimension to ensure the global consistency of denoising decisions. Channel attention is responsible for evaluating the importance of the features and screening the information in the channel dimension to ensure that the features used for decision making are optimal. Therefore, it can preserve or even enhance the useful structural information of the SAR images to the maximum extent and generate high-quality denoising results while strongly suppressing speckle noise.

In order to obtain valuable ship features, we first perform a preliminary feature extraction on the image before it enters the CID module. We choose $1 \times 1$ convolutions instead of $3 \times 3$ convolutions due to the computational complexity of high-resolution images. The details of the structure of the enhanced CID module are shown in Fig.~\ref{CID}(a), and the output mode is the shortest path because the feature channel is converted from $3$ to $64$. To be specific, for the input feature $I$, the output feature $I^{'}$ after the channel-independent denoising block can be formulated as:
\begin{equation}
    I^{'} =BN(PConv(CID(I))), CID=MLP(DConv7(\cdot)).
\end{equation}
Where BN is a 2D BatchNorm. $\mathrm{DConv7}$ is a depth-wise convolution with $7 \times 7$ kernels. $\mathrm{MLP}$ is implemented by two point-wise convolutional layers and a $\mathrm{GELU}$ \cite{R6} non-linearity function. 

These easy computations make up the CID module, and channel attention ensures that the relationship between features across different channels can be effectively avoided in terms of feature extraction. Adding a denoising module to the model enables the model to pay more attention to the ship information in SAR images and prevent missed detections caused by image noise affecting ships with fewer pixels.

(3) PPA: In small target detection tasks on the ship, {YOLOv8n} adds Spatial Pyramid Pooling Fast (SPPF) to the end of the backbone extraction network to enhance feature extraction. However, it is easy to lose key information in the process of small target down-sampling. To better address this challenge, Xu et al. \cite{M3} designed a parallelized patch-aware attention module, which consists of three key points: (1) multi-branch feature extraction; (2) feature selection; (3) feature fusion and attention.

\begin{figure*}[t]
    \centering
    \includegraphics[width=0.95\linewidth]{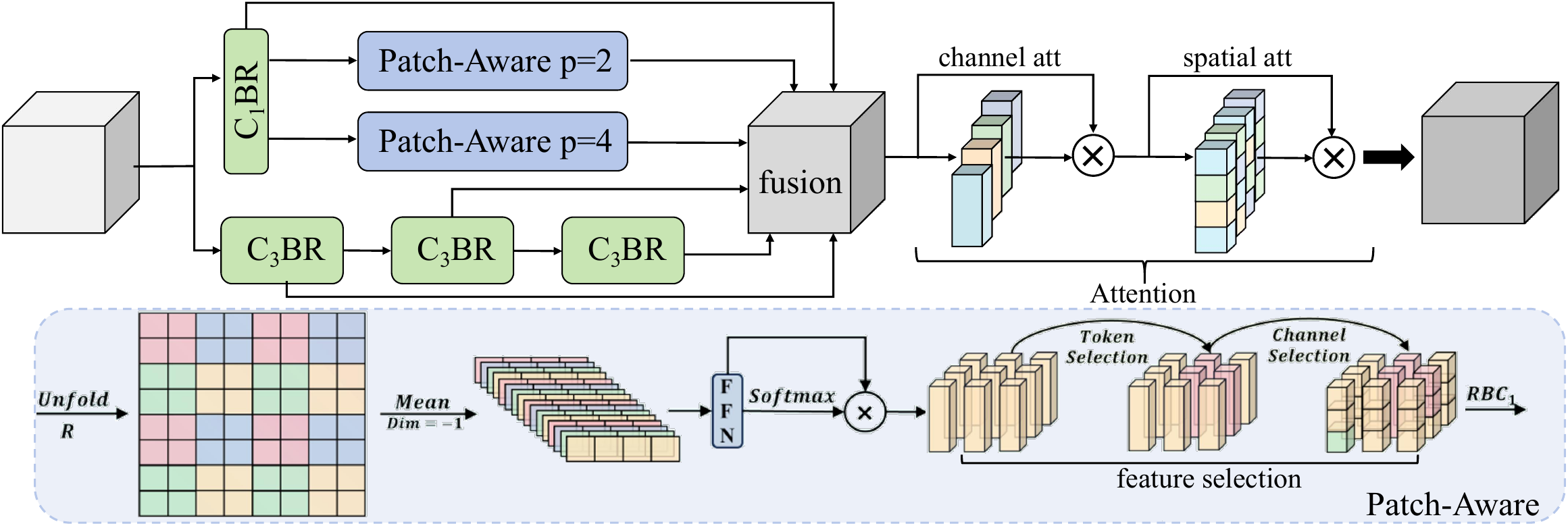}
    \caption{Structure of the Parallelized Patch-Aware Attention Module. It is composed of two core components: a multi-branch fusion unit and an attention mechanism. The multi-branch fusion unit integrates two parallel operations: patch-aware convolution and concatenated convolution. In the patch-aware convolution, the hyperparameter pis configured to 2 and 4, corresponding to local and global receptive field branches, respectively.}
    \label{PPA}
\end{figure*}

For this purpose, we introduce PPA (Fig.~\ref{PPA}) after the SPPF at the end of the feature extraction network. With the application of the PPA attention mechanism, the new characterization learning enhancement structure, SPAF is introduced to fully use the distinctive features of ship targets. This module forces the network to extract features of different scales and levels through multiple parallel branches: local, global and tandem convolution branches, thus improving the accuracy of small object detection. Specifically, given the input feature tensor $X \in R^{H^{'} \times W^{'} \times C } $, it is first sent to point-wise convolution for preliminary feature extraction to obtain$X^{'} \in R^{H^{'} \times W^{'} \times C } $. And then, they are sent to Patch-Aware ({p=2 \& p=4}) to compute $X_{local} \in R^{H^{'} \times W^{'} \times C^{'} } $, $X_{global} \in R^{H^{'} \times W^{'} \times C^{'} } $, respectively. In addition, the input features are also sent to the concatenated branch at the same time to calculate the features $X^{1} \in R^{H^{'} \times W^{'} \times C}$, $X^{2} \in R^{H^{'} \times W^{'} \times C}$, $X^{3} \in R^{H^{'} \times W^{'} \times C}$ as follows:
\begin{equation}
    X_{(\cdot)} =PA_p(C_1BR(X)).
\end{equation}
Where $X_{(\cdot)}$ denotes $X_{local}$ or $X_{global}$, CBR respectively represents convolution, $\mathrm{BN}$ and $\mathrm{ReLU}$ activation function, while the lower number indicates the kernel size. $PA_p$ is the Patch-Aware module, $p$ is the patch size parameter. In particular, $\mathrm{PA}$ begins by unfolding and reshaping the features to divide them into spatially continuous feature blocks. Then, a channel-wise mean operation is applied to compress the feature depth. The compressed features are passed through an $\mathrm{FFN}$ and activated to produce a probability distribution in the spatial domain. This distribution is utilized to scale the original $\mathrm{FFN}$ output, thereby adjusting their importance based on spatial context. After that, to select the suitable features for the task, the feature selection is performed according to \cite{M4}, and the advanced token selection($f_{select-t}$) is selected followed by the channel selection($f_{select-c}$). For a more concise expression, we divide the specific operation of $\mathrm{PA}$ into two parts($PA_1$ and $PA_2$), as shown in the following:

\begin{equation}
    PA_1=FFN(f_{Mean_{dim=-1}}(f_{Unfold(X^{'})})).
\end{equation}
\begin{equation}
    PA_2=f_{select-t}(f_{select-c}(Softmax(PA_1) \otimes PA_1)).
\end{equation}
Where $f_{Unfold(\cdot)}$ is the unfold operation, $f_{Mean_{dim=-1}}$ is the mean operation for the last dimension, $\mathrm{FFN}$ is the same linear computations followed by \cite{ffn} and $\mathrm{Softmax}$ represent the normalized exponential function. 

In addition, to express the relationship from $X^1$ to $X^3$ more clearly, we use the following specific recursive formula:
\begin{equation}
    PA_2=f_{select-t}(f_{select-c}(Softmax(PA_1) \otimes PA_1)).
\end{equation}
\begin{equation}
   X^n=
\begin{cases}
  & C_3BR(X^{n-1}), n>1. \\
  & C_3BR(X), n=1.
\end{cases}
\end{equation}
Among them, as before, $C_3BR$ represents the convolution module that combines $BN$ and $ReLU$ with the convolution whose kernel size is $3$. In total, there are six outputs for the extraction of multiple branch features, which are $X^{'}, X^1, X^2, X^3, X_{local}$ and $X_{global}$. To fuse the features of different branches, we sum them and serially convolution to obtain the output $X^{branch}$. Following multi-branch feature extraction and selection, PPA performs adaptive feature enhancement through a cascaded attention mechanism comprising a one-dimensional channel attention module \cite{M5} and a two-dimensional spatial attention module \cite{M6}. Specifically, the branch feature $X^{branch}$ is first modulated by the channel attention mechanism($A_C$) to re-weight inter-channel responses, and is subsequently refined by the spatial attention mechanism($A_S$) to emphasize salient regions. The overall procedure is summarized as follows:
\begin{equation}
    X_c=A_C(X^{branch}) \otimes X^{branch}, X_S=A_S(A_C) \otimes A_C.
\end{equation}
Where $\otimes$ denotes element-wise multiplication, $X_c$ and $X_S$ represent features after channel and spatial attention. Finally, the convolution module is performed again to obtain the final output of PPA $X_{PPA}$. 
\begin{equation}
    X_{PPA}=\delta (\mathcal{B}(dropout(X_S))).
\end{equation}
$\delta (\cdot)$ and $\mathcal{B}(\cdot)$ represent the Rectified Linear Unit ($ReLU)$ and Batch Normalization ($BN$), respectively.

(4) NWDLoss \cite{M7}: The bounding box regression involves the precise localization of the object within the confines of the detected target, which is of significant importance in ship detection. Despite the size of most SAR photos, some of them contain only a few ship pixels. In addition, the most advanced detectors do not produce satisfactory results on tiny objects due to the lack of appearance information. 

Follow the following facts: (1) Most real objects are not strictly rectangles. Their bounding boxes often contain some background pixels, in which the foreground pixels and background pixels are, respectively, concentrated at the center and the boundary of the bounding box; (2) Different pixels in the bounding box have different contributions to the prediction. The center pixel of the bounding box has the highest weight, and the importance of the pixel decreases from the center to the boundary. We introduce the Wasserstein Distance to evaluate the distance and similarity between tiny objects. Different from the traditional location-based judgments such as IoU and its improved GIoU \cite{M9}, DIoU and CIoU \cite{M8}, NWDLoss first models the bounding box as a two-dimensional Gaussian distribution. Then, the normalized Wasserstein distance is calculated to measure the similarity of the Gaussian distribution, which can not only measure the distribution similarity between object boxes with no overlap or negligible overlap, but also reduce the sensitivity of the loss to objects of different scales, making it more suitable for measuring the similarity between small objects. The equation below can be used to explain the loss function:
\begin{equation}
    \mathcal{L}_{NWD}=1-NWD(\mathcal{N}_a, \mathcal{N}_b).
\end{equation}

It includes four parts:
(1) Change the discrete bounding box function to the continuous inscribed elliptic function, defined as follows:
\begin{equation}
    \frac{(x-\mu_x )^{2}}{\sigma _{x}^{2} } +\frac{(y-\mu_y )^{2}}{\sigma _{y}^{2}} =1.
\end{equation}
Where $(\mu_x,\mu_y)$ represents the center coordinates of the ellipse for the bounding box $R =$ $(c_x,c_y,w,h)$ that $(c_x,c_y)$, $w$ and $h$ denote the center coordinates, width, and height, respectively. $\sigma _{x}$ and $\sigma _{y}$ are the lengths of the semi-axes along the $x$ and $y$ axes. Accordingly, $\mu_{x}=c_x, \mu_{y}=c_y, \sigma_{x}=\frac{w}{2}, \sigma_{y}=\frac{h}{2}$.

(2) Model the bounding box into two-dimensional (2D) Gaussian distribution $\mathcal{N}(\mu,\sum)$, defined as follows:
\begin{equation}
    \mu =
\begin{bmatrix}
 c_x \\
 c_y
\end{bmatrix},
\sum =\begin{bmatrix}
 \frac{w^2}{4} & 0 \\
  0 & \frac{h^2}{4}
\end{bmatrix}.
\end{equation}
Where $[\cdot]$ denotes a two-dimensional matrix, $\mu$ and $\sum$ are the mean vector and the Gaussian distribution covariance matrix. More theoretical derivations are provided in \cite{M7}. Here, the similarity between the bounding boxes $A$ and $B$ can be converted into the distribution distance between the two Gaussian distributions.

(3) Calculate the distance between two distributions $\mathcal{N}_a$ and $\mathcal{N}_b$ based on the Wasserstein distance, defined as follows:
\begin{equation}
    W_{2}^{2}(\mathcal{N}_a,\mathcal{N}_b)=\left \| ([c_{x_a},c_{y_a},\frac{w_a}{2},\frac{h_a}{2}]^T,[c_{x_b},c_{y_b},\frac{w_b}{2},\frac{h_b}{2}]^T) \right \|_{2}^{2} .
\end{equation}
Where $\left \| \cdot  \right \|_{2}^{2} $ is the Frobenius Norm when $F=2$. $c_x, c_y, \frac{w}{2}$ and $\frac{h}{2}$ come from bounding boxes $A=$$(c_{x_a},c_{y_a},w_a,h_a) $ and $B=$$(c_{x_b},c_{y_b},w_b,h_b) $ that follow the Gaussian distribution.

(4) Obtain the similarity measure from the normalize Wasserstein distance, defined as follows:
\begin{equation}
    NWD(\mathcal{N}_a,\mathcal{N}_b)=exp(-\frac{\sqrt{W_{2}^{2}(\mathcal{N}_a),\mathcal{N}_b}}{c}) .
\end{equation}
Where $c$ is a constant closely related to the dataset, $exp(\cdot)$ stands for exponential operation. According to the original paper, this conversion step is due to $W_2^2(\mathcal{N}_a, \mathcal{N}_b)$ being a distance measure and cannot be directly used as a similarity measure (the values between 0 and 1 are taken as IoU). 
\section{Experiment and results}
\subsection{Experimental Environment}
In this article, all experiments are performed with the Ubuntu 22.04 operating system, using Pytorch 2.1.0 and Python 3.10 for training, CUDA 12.1 is used to speed up the calculation, and the processor on the computer is {14 vCPU Intel(R) Xeon(R) Gold 6348 CPU @ 2.60GHz}, with 100 GB memory and an A800 graphics card.
\subsection{Datasets}
To assess the reliable performance of the proposed method in the model in various datasets, experiments are conducted in two publicly available datasets: the SAR Ship Detection Dataset ({SSDD}) \cite{E1} and the High-Resolution SAR Images Dataset ({HRSID}) \cite{E2} . In addition to the typical views of ships in the far-off sea, both datasets include a wide range of scenarios such as inshore, ports, and islands. The two datasets will then be introduced, and further details are provided below.
\begin{figure*}[t]
    \centering
    \includegraphics[width=0.95\linewidth]{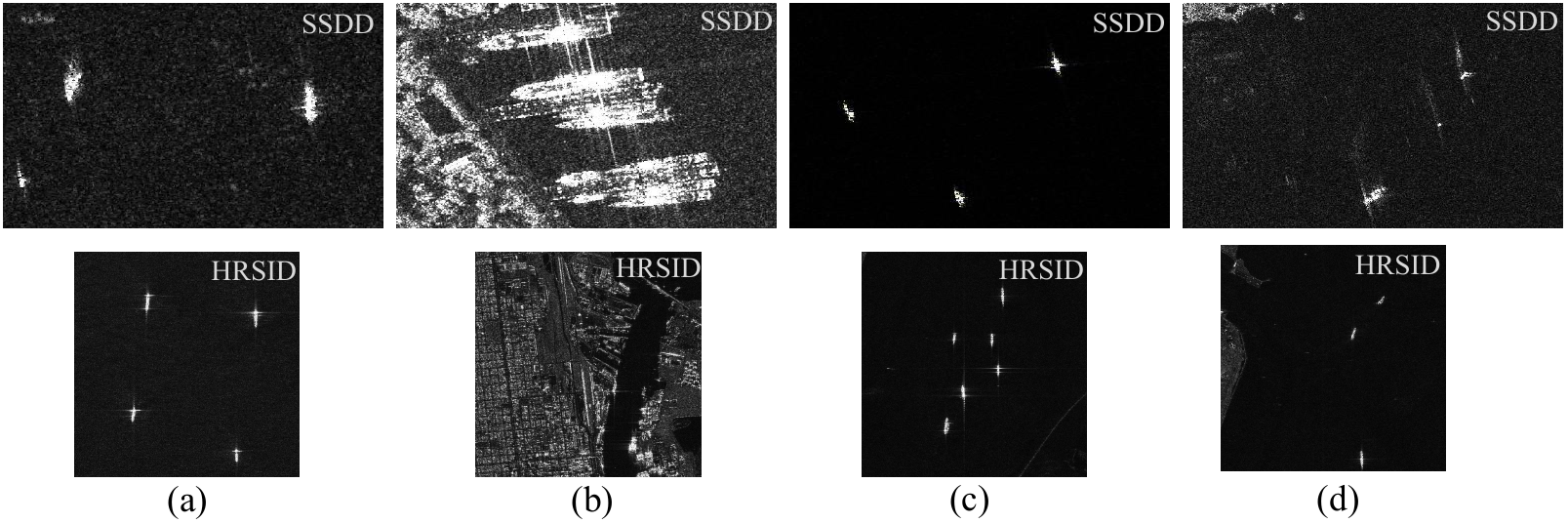}
    \caption{Samples of SSDD and HRSID dataset. (a) Multiscale ship samples, (b) inshore ship samples, (c) small ship samples, (d) Offshore ship samples.}
    \label{dataset}
\end{figure*}

\noindent\textbf{SSDD:} The {SSDD} dataset is presented by \cite{E1}. It contains 1160 images, among which 928 images are used for training and 232 image are used for testing, and 2456 ships of different sizes collected from Sentinel-1, TerraSAR, and RadarSat-2, which provides the raw data, with image resolutions between 1 and 15 megapixels. Each image has an approximate size of 600 pixels in length and width, and the imaging scenes include complex scenes such as docks and the near shore environment, and simple scenes such as the sea surface. Each SAR image contains a different number, size, and category of ship objects. In this dataset, {60.2\%}, {36.8\%}, and {3.0\%} of all ships are made up of small, medium and large ships, respectively. In addition, the image labels of the dataset are manually annotated by referring to the {PASCAL VOC} dataset format. Sample images from the {Official-SSDD} dataset are illustrated in Fig.~\ref{dataset}.

\noindent\textbf{HRSID:} The raw data from the {HRSID} dataset is presented by \cite{E2}. It contains 5604 images, of which 3922 images are used for training and 1682 images are used for testing, and 16951 ships are obtained from three satellite sensors, TerraSAR-X, TanDEM-X, and Sentinel-B, and provides a panoramic view of SAR images with a {25\%} overlap rate, with image resolutions between 1 and 5 m. Each image is sized at 800 $\times$ 800 pixels. The imaging scenes include complex scenes such as ports, docks and the near shore environment, and simple scenes such as the sea surface. Each SAR image contains a different number of ship targets of different sizes. In this dataset, {54.5\%}, {43.5\%}, and {2.0\%} of all ships, respectively, are small, medium and large ships. In addition, the image labels of the dataset are manually annotated by referring to the {MS COCO} dataset format. The sample images extracted from the {HRSID} dataset are shown in Fig. ~\ref{dataset}.
\subsection{Performance Metrics}
To evaluate the effectiveness of the {RSTNet} method, we employ the same three key performance evaluation metrics as outlined in the Ultralytics office and MS COCO \cite{E3}: precision (P), recall (R) and average precision (AP). These metrics consist of the three components: true positive (TP), false positive (FP) and false negative (FN). TP refers to the number of correctly matched ships under the special IoU threshold, while FP corresponds to the number of incorrectly matched ships. FN denotes the number of incorrectly matched as background. The matchematically expressed of P, R and AP as:
\begin{equation}
    P=\frac{TP}{TP+FP}
\end{equation}
\begin{equation}
    R=\frac{TP}{TP+FN}
\end{equation}
\begin{equation}
    AP=\int_{0}^{1}P(R)dR 
\end{equation}

\noindent\textbf{IoU-based Average Precision:} Localization performance is quantified by AP computed under varying IoU thresholds. Specifically, $AP_{50}$ is obtained by treating detection with IoU $\ge 0.50$ against ground-truth boxes, while $AP_{75}$ employs a stricter IoU criterion $\ge 0.75$. To reflect robustness in different localization stringencies, the primary $AP_{50-95}$ is defined as thresholds ranging from 0.50 to 0.95 in increments of 0.05. 

\noindent\textbf{Scale-aware Average Precision:} In addition to localization, real world applications often demand reliable detection of objects spanning diverse scales. We therefore partition samples into three volumetric subsets according to their pixel area: (1)$AP_S$ refers to the AP calculated for objects with a volume smaller than $32^2$ pixels; (2) $AP_M$ refers to the AP for objects with a volume between $32^2$ and $96^2$ pixels; (3) $AP_L$ refers to the AP for objects with a volume greater than $96^2$ pixels.

\subsection{Experiment Comparison and Analysis}
To evaluate the performance of the proposed {RSTNet}, we conduct a comprehensive set of experiments involving systematic modifications to the activation functions and architectural configurations of the baseline models. These results not only demonstrate the efficacy of our approach, but also further guide the optimization of the model.

In this section, all experiments are conducted on the {SSDD} dataset under a unified configuration to ensure consistency across evaluations: training continues for 200 epochs using the SGD optimizer, with an initial learning rate of 0.01, a batch size of 64, and an input resolution of 640 $\times$ 640 pixels. The results are evaluated by Ultralytics office.

\noindent\textbf{(1) Effect of different components:} To assess the impact of various improved components in {RSTNet}, we conduct comparative experiments to compare the advantages, disadvantages, and effects of various improvement strategies. Quantitative results are summarized in Tab.~\ref{tab different components}.

\begin{table}[]
\centering
\caption{Ablation experiments for different components.}
\begin{tabular}{clllll}
\hline
NO. & Model             & \multicolumn{1}{r}{P} & \multicolumn{1}{r}{R} & \multicolumn{1}{r}{mAP@0.5} & \multicolumn{1}{r}{mAP@0.5:0.95} \\ \hline
1   & YOLOv8            & 0.938                 & 0.936                 & 0.976                                              & 0.693                                                   \\
2   & CID               & 0.964                 & 0.93                  & 0.976                                              & 0.721                                                   \\
3   & CID\&NWDLoss      & 0.961                 & \textbf{0.953}        & 0.984                                              & 0.727                                                   \\
4   & CID\&NWDLoss\&PPA & \textbf{0.97}         & 0.951                 & \textbf{0.989}                                     & \textbf{0.739}                                          \\ \hline

\end{tabular}
\label{tab different components}
\end{table}

Firstly, we introduce the CID module before the original {YOLOv8} feature extraction stage. Tab.~\ref{tab different components} shows that the addition of this module significantly improves accuracy (P) and mAP@0.5:0.95 of SAR ship target detection. The main reason is that the traditional feature extraction method uses an equal treatment strategy for each channel information and directly stacks a large number of small kernel convolutions. Although it can extract high-level semantic features, it will weaken the representation strength of important channels and limit the receptive field to local areas, making it difficult to effectively capture global context information. It significantly expands the perception field while retaining the ability of local detail feature extraction, so that the model can obtain multi-scale context cues, and assigns different channels with differentiated importance by means of the channel-level weight allocation mechanism, so that the network can more specifically strengthen task-related features and suppress redundant or noise responses. Furthermore, the results of experiment NO.3 show that the use of the NWDLoss function can significantly improve the detection indicators, except for P. This is because when predicting the regression of the boundary box, NWDLoss reduces the loss sensitivity to the target scale by redefining the measure of similarity of the target box distribution, thereby improving the accuracy of the regression and the overall detection performance. Finally, after adding the attention module PPA, the model's mAP@0.5 increases to {98.9\%} (relative increase of {1.3\%}) and mAP@0.5:0.95 increases to {73.9\%} (relative increase of{ 4.6\%}). It shows that the PPA module can effectively alleviate the problem of key information loss caused by multiple down-sampling of {YOLOv8}, thereby reducing the missed detection phenomenon. In summary, the method proposed in this paper makes systematic improvements in three aspects of feature extraction, loss function design, and attention enhancement, and each module complements each other. The model can obtain a more comprehensive and robust representation of features during the training process, so it is more suitable for SAR ship detection tasks.

\noindent\textbf{(2) Effect of PPA:} In addition to the ablation studies on individual components, we further evaluate the effectiveness of the proposed PPA module through a comparative analysis with three alternative attention mechanisms: squeeze-and-excitation (SE) \cite{E4}, coordinate attention (CA) \cite{E5} and convolutional block attention module (CBAM) \cite{E6}. All experiments incorporate at identical network locations and detail experimental results are presented in Tab.~\ref{tab PPA}.
\begin{table}[]
\centering
\caption{Comparative results of different attention mechanism models.}
\begin{tabular}{clrrrr}
\hline
NO. & Model & P              & R              & mAP@0.5 & mAP@0.5:0.95 \\ \hline
1   & CA \cite{E5}    & 0.956          & 0.96           & 0.984                          & 0.721                               \\
2   & CBAM \cite{E6}  & 0.945          & \textbf{0.971} & \textbf{0.985}                 & 0.716                               \\
3   & SE \cite{E4}    & 0.959          & 0.954          & \textbf{0.985}                 & 0.723                               \\
4   & Ours   & \textbf{0.966} & 0.939          & 0.983                          & \textbf{0.735}                      \\ \hline
\end{tabular}
\label{tab PPA}
\end{table}
Performance analysis demonstrates that the proposed PPA module achieves superior results, achieving a precision of {96.6\%}, a recall of {93.9\%}, an mAP@0.5 of {98.3\%}, and an mAP@0.5:0.95 of {73.5\%}. Compared with the CA, CBAM and SE models, PPA yields absolute gains in precision of {1.0\%}, {2.1\%} and {0.7\%}, respectively, and improvements in mAP@0.5:0.95 of {1.4\%}, {1.9\%} and {1.2\%}, respectively. These results indicate that PPA delivers substantial advances in detection performance over the evaluated attention mechanisms, underscoring its effectiveness in enhancing both localization accuracy and overall detection quality.

\begin{figure*}[t]
    \centering
    \includegraphics[width=0.55\linewidth]{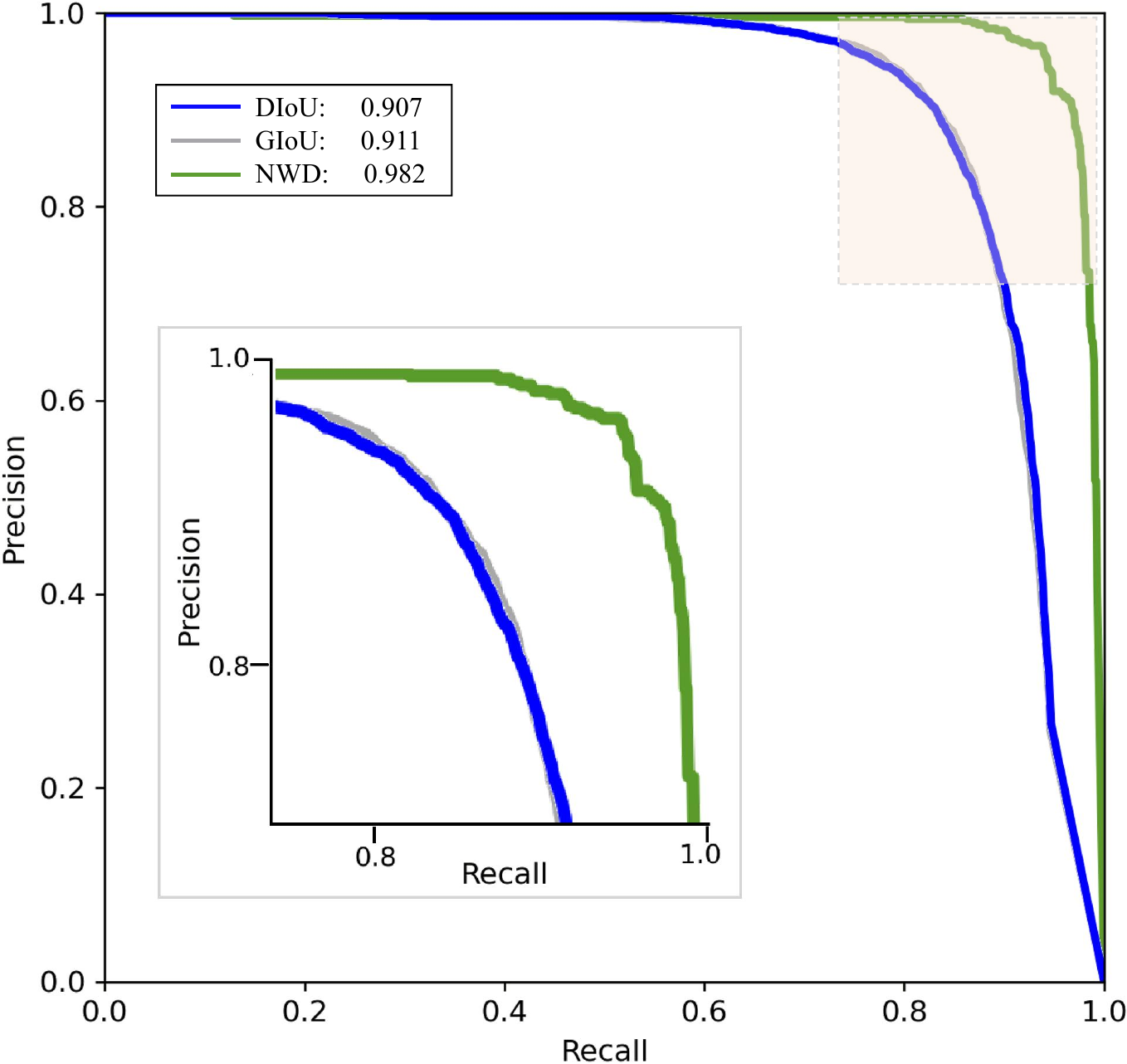}
    \caption{AP curves of different BBR loss function.}
    \label{PR curves}
\end{figure*}

\noindent\textbf{(3) Effect of NWDLoss:} Following the enhancements to the backbone and feature fusion modules, this section presents a comparative analysis of bounding box regression (BBR) loss functions. We  evaluate three representative losses DIoU \cite{E7}, GIoU \cite{E8}, and NWDLoss against the baseline model. As shown in Tab.~\ref{tab loss}, NWDLoss yields an mAP@0.5:0.95 of {71.6\%}, representing absolute improvements of {6.6\%} and {6.2\%} over DIoU and GIoU, respectively. These results demonstrate that NWDLoss achieves superior detection accuracy, indicating its effectiveness in enhancing the localization performance for the SAR detection task.
\begin{table}[]
\centering
\caption{Ablation experiments for different components.}
\begin{tabular}{clrrrr}
\hline
NO. & Model & P              & R             & mAP@0.5 & mAP@0.5:0.95 \\ \hline
1   & DIoU \cite{E7}  & 0.908          & 0.816         & 0.907                          & 0.65                                \\
2   & GIoU \cite{E8} & 0.911          & 0.822         & 0.911                          & 0.654                               \\
3   & Ours   & \textbf{0.963} & \textbf{0.94} & \textbf{0.982}                 & \textbf{0.716}                      \\ \hline
\end{tabular}
\label{tab loss}
\end{table}
To assess the performance of different BBR loss functions in terms of precision recall characteristics, the precision recall curves of DIoU, GIoU, and NWDLoss functions are studied in this section,as shown in Fig.~\ref{PR curves}. The findings suggest that the NWDLoss function achieves the maximum area enclosed by the X and Y axes at different recall levels, with an average precision of 0.982, outperforming DIoU (0.907) and GIoU (0.911). This indicates that the NWDLoss function leads to more accurate object detection results. Specifically, in high recall regions (e.g., recall $>$ 0.8), the NWDLoss function maintains a significantly higher precision compared to DIoU and GIoU, as highlighted in the inset figure. This underlines the superior detection performance achieved by NWDLoss in terms of precision and its ability to handle object detection tasks more effectively in complex scenarios.

\noindent\textbf{(4) Comparison with other ship detection models:} To evaluate the robust detection capability of RSTNet on SAR images, we perform a comparative analysis against nine widely adopted SAR ship detection models: {C-AFBiFPN}, {YOLOv10}, {DiffusionDet}, {YOLOv8}, {YOLOv5}, {FCOS}, {RetinaNet}, {SSD}, and {Faster R-CNN}, under identical experimental settings and hyperparameter configurations. Two publicly available datasets, {SSDD} and {HRSID}, are used for performance assessment. The comparative results(according to MS COCO) are summarized in Tab.~\ref{tab HRSID} and Tab.~\ref{tab SSDD}, where a refined performance comparison between {RSTNet} and the competing models on the {SSDD} dataset is illustrated in Fig.~\ref{SSDD_result}, and a corresponding comparison on the {HRSID} dataset is presented in Fig.~\ref{HRSID_result}.

Tab.~\ref{tab HRSID} presents a performance comparison of the same models in the {HRSID} dataset. In this dataset, {RSTNet} achieves the highest mAP@0.5 ({88.9\%}), mAP@0.75 ({73\%}), mAP@0.5:0.95 ({64.5\%}), $mAP_S$ ({52.6\%}) and $mAP_M$ ({80.4\%}) exceeding the other models. This result highlights the superior performance of {RSTNet} in detecting ships in the {HRSID} dataset.

\begin{table*}[]
\centering
\caption{Performance evaluation of various models on the HRSID dataset. Sorted by mAP@0.5 positive order.}
\resizebox{0.9\textwidth}{!}{
\begin{tabular}{lclllllll}
\hline
\multicolumn{1}{l}{Model} & Year        & Dataset & \multicolumn{1}{r}{mAP@0.5} & \multicolumn{1}{r}{mAP@0.75} & \multicolumn{1}{r}{mAP@0.5:0.95} & \multicolumn{1}{r}{$mAP_S$} & \multicolumn{1}{r}{$mAP_M$} & \multicolumn{1}{r}{$mAP_L$} \\ \hline
Faster Rcnn \cite{FasterRcnn} & \begin{tabular}[c]{@{}c@{}}2015\\ TPAMI\end{tabular}   & HRSID   & 0.762                                                    & 0.619                                                     & 0.540                                                          & 0.382                    & 0.760                     & 0.494                    \\
C-AFBiFPN \cite{C-AFBiFPN}    & \begin{tabular}[c]{@{}c@{}}2025\\ arxiv\end{tabular}  & HRSID   & 0.763                                                    & 0.636                                                     & 0.552                                                         & 0.404                    & 0.766                    & 0.416                    \\
RetinaNet \cite{RetinaNet}   & \begin{tabular}[c]{@{}c@{}}2017\\ TPAMI\end{tabular}   & HRSID   & 0.771                                                    & 0.568                                                     & 0.522                                                         & 0.354                    & 0.746                    & 0.433                    \\
DiffusionDet \cite{diffusionDet} & \begin{tabular}[c]{@{}c@{}}2023\\ CVPR\end{tabular}   & HRSID   & 0.840                                                     & 0.678                                                     & 0.588                                                         & 0.468                    & 0.755                    & 0.523                    \\

FCOS  \cite{FCOS}       & \begin{tabular}[c]{@{}c@{}}2019\\ CVPR\end{tabular}   & HRSID   & 0.858                                                    & 0.660                                                      & 0.599                                                         & 0.463                    & 0.778                    & 0.50                      \\

SSD \cite{SSD}          & \begin{tabular}[c]{@{}c@{}}2016\\ ECCV\end{tabular}   & HRSID   & 0.872                                                    & 0.654                                                     & 0.586                                                         & 0.452                    & 0.777                    & 0.489                    \\
YOLOv10 \cite{yolov10}      & \begin{tabular}[c]{@{}c@{}}2024\\ arxiv\end{tabular}  & HRSID   & 0.875                                                    & 0.720                                                      & 0.629                                                         & 0.513                    & 0.786                    & 0.514                    \\

YOLOv5 \cite{yolov5}      & \begin{tabular}[c]{@{}c@{}}2020\\ github\end{tabular}    & HRSID   & 0.882                                                    & 0.710                                                      & 0.627                                                         & 0.507                    & 0.789                    & 0.521                    \\
YOLOv8 \cite{yolov8}      & \begin{tabular}[c]{@{}c@{}}2023\\ github\end{tabular} & HRSID   & 0.887                                                    & 0.714                                                     & 0.628                                                         & 0.506                    & 0.791                    & \textbf{0.568}           \\
\multicolumn{2}{c}{\textbf{Ours}}                                    & HRSID   & \textbf{0.889}                                           & \textbf{0.730}                                             & \textbf{0.645}                                                & \textbf{0.526}           & \textbf{0.804}           & 0.562                    \\ \hline
\multicolumn{9}{l}{The bold values indicate the maximun value of the evaluation.}                                                             
\end{tabular}
} 
\label{tab HRSID}
\end{table*}
Tab.~\ref{tab SSDD} provides a performance comparison of various models in the {SSDD} dataset. The results reveal that the {RSTNet} model exhibits the highest mAP@0.5 ({97.3\%}), while the {C-AFBiFPN} shows the highest mAP@0.75 ({77.6\%}). {RSTNet} also achieves the best performance of mAP@0.5:0.95, with a value of {66.5\%}. In particular, it also exhibits superior performance in terms of $mAP_M$ and $mAP_L$. 

The results presented in Tab.~\ref{tab HRSID} and Tab.~\ref{tab SSDD} demonstrate the superior performance of our proposed method {HRSID} on and {SSDD} datasets, consistently outperforming existing models. This advantage can be primarily ascribed to two key design elements: an improved denoising pipeline and a tailored attention mechanism, which jointly facilitate the efficient extraction of discriminative characteristics of the ship and thus strengthen the predictive accuracy of the model. Furthermore, the adoption of the NWDLoss function improves the precision of target box regression, contributing to the enhanced generalization capability and robustness in complex scenarios.

\begin{table*}[]
\centering
\caption{Performance evaluation of various models on the SSDD dataset. Sorted by mAP@0.5 positive order.}
\resizebox{0.9\textwidth}{!}{
\begin{tabular}{lclllllll}
\hline
\multicolumn{1}{l}{Model} & Year                                                  & Dataset & \multicolumn{1}{r}{mAP@0.5} & \multicolumn{1}{r}{mAP@0.75} & \multicolumn{1}{r}{mAP@0.5:0.95} & \multicolumn{1}{r}{$mAP_S$} & \multicolumn{1}{r}{$mAP_M$} & \multicolumn{1}{r}{$mAP_L$} \\ \hline
RetinaNet \cite{RetinaNet}                & \begin{tabular}[c]{@{}c@{}}2017\\ TPAMI\end{tabular}   & SSDD    & 0.938                                              & 0.679                                               & 0.595                                                         & 0.557                    & 0.672                    & 0.621                    \\
Faster Rcnn \cite{FasterRcnn}              & \begin{tabular}[c]{@{}c@{}}2015\\ TPAMI\end{tabular}   & SSDD    & 0.948                                              & 0.698                                               & 0.61                                                          & 0.581                    & 0.668                    & 0.567                    \\
YOLOv5 \cite{yolov5}                   & \begin{tabular}[c]{@{}c@{}}2020\\ github\end{tabular}    & SSDD    & 0.950                                              & 0.693                                               & 0.616                                                         & 0.571                    & 0.709                    & 0.560                    \\
YOLOv10 \cite{yolov10}                  & \begin{tabular}[c]{@{}c@{}}2024\\ arxiv\end{tabular}  & SSDD    & 0.951                                              & 0.703                                               & 0.608                                                         & 0.571                    & 0.693                    & 0.564                    \\
SSD \cite{SSD}                      & \begin{tabular}[c]{@{}c@{}}2016\\ ECCV\end{tabular}   & SSDD    & 0.951                                              & 0.731                                               & 0.617                                                         & 0.569                    & 0.692                    & 0.693                    \\
FCOS \cite{FCOS}                     & \begin{tabular}[c]{@{}c@{}}2019\\ CVPR\end{tabular}   & SSDD    & 0.959                                              & 0.741                                               & 0.632                                                         & 0.587                    & 0.711                    & 0.736                    \\
YOLOv8 \cite{yolov8}                   & \begin{tabular}[c]{@{}c@{}}2023\\ github\end{tabular} & SSDD    & 0.959                                              & 0.725                                               & 0.632                                                         & 0.588                    & 0.722                    & 0.584                    \\
C-AFBiFPN \cite{C-AFBiFPN}                & \begin{tabular}[c]{@{}c@{}}2025\\ arxiv\end{tabular}  & SSDD    & 0.961                                              & \textbf{0.776}                                      & 0.66                                                          & \textbf{0.643}           & 0.695                    & 0.625                    \\

DiffusionDet \cite{diffusionDet}             & \begin{tabular}[c]{@{}c@{}}2023\\ CVPR\end{tabular}   & SSDD    & 0.967                                              & 0.735                                               & 0.634                                                         & 0.571                    & 0.734                    & 0.726                    \\

\multicolumn{2}{c}{\textbf{Ours}}                                                 & SSDD    & \textbf{0.973}                                     & 0.772                                               & \textbf{0.665}                                                & 0.607                    & \textbf{0.75}            & \textbf{0.749}           \\ \hline
\multicolumn{9}{l}{The bold values indicate the maximun value of the evaluation.} 
\end{tabular} }  
\label{tab SSDD}
\end{table*}

To further estimate the performance variations of each model in different datasets, a direct comparison of the results in Tab.~\ref{tab SSDD} and Tab.~\ref{tab HRSID} enable the assessment of the accuracy of the same model’s in different benchmarks. The data indicate that, across all nine evaluated models, the performance metrics including mAP@0.5 and mAP@0.75 on the {SSDD} dataset are consistently higher than the corresponding values on the {HRSID} dataset. This discrepancy can be largely ascribed to the high quality of {SSDD} annotation, which ensures precise and consistent ship labeling. In addition, {SSDD} typically exhibits a balanced distribution of positive and negative samples, with comparable numbers of ship and non-ship instances; such balance facilitates more stable model training and consequently enhances detection performance. In contrast, {HRSID} poses greater challenges, as it consists of high resolution SAR imagery that contains fine structural details along with inherent noise, thereby increasing the difficulty of accurate localization and classification.

\begin{figure*}[t]
    \centering
    \includegraphics[width=0.95\linewidth]{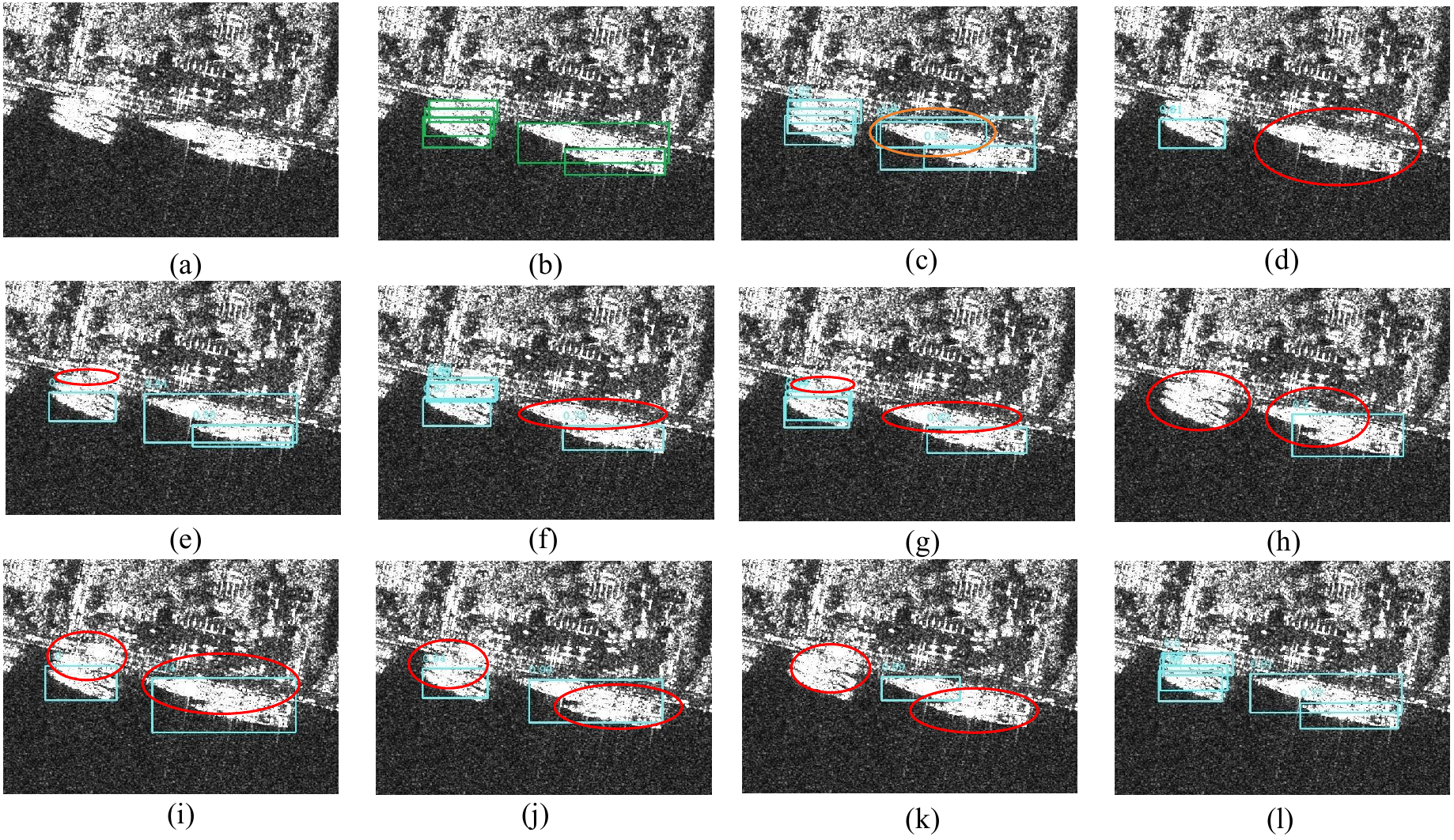}
    \caption{Revised comparison chart: Performance of RSTNet and other models on the SSDD dataset. (a) Original image. (b) Ground Truth. (c) C-AFBiFPN visualization results. (d) YOLOv10 visualization results. (e) DiffusionDet visualization results. (f) YOLOv8 visualization results. (g) YOLOv5 visualization results. (h) FCOS visualization results. (i)  RetaiNet visualization results. (j) SSD visualization results. (k) Faster Rcnn visualization results. (l) RSTNet visualization results. The Green rectangles indicate GT, blue rectangles indicate predicted results, and red ellipses indicate missed detections.}
    \label{SSDD result}
\end{figure*}

Fig.~\ref{SSDD result} presents a comparative visualization of the detection performance between {RSTNet} and the other nine models on the {SSDD} dataset. In Fig.~\ref{SSDD result} (a) and (b), the original SAR image and its corresponding ground truth are shown, with five inshore ship targets annotated. Fig.~\ref{SSDD result} (c)–(l) display the detection results of {C-AFBiFPN}, {YOLOv10}, {DiffusionDet}, {YOLOv8}, {YOLOv5}, {FCOS}, {RetinaNet}, {SSD}, {Faster R-CNN} and {RSTNet}, respectively. All five ships are located in nearshore regions. Visual inspection reveals that most models exhibit occasional missed detections (marked with red ellipses) along with correct identifications (marked with blue rectangles). In particular, the {SSDD} dataset demonstrates better adaptability and recognition accuracy compared to the {HRSID} dataset, as reflected in these visualizations. Specifically, {RetinaNet} correctly detects only one ship and misses four, exhibiting results identical to {SSD}. Three models: {YOLOv10}, {Faster R-CNN}, and {FCOS} show the lowest detection accuracy, frequently misclassifying nearshore ships as background. {C-AFBiFPN} accurately localizes all five ships, but yields one false positive (FP). {C-AFBiFPN}, {DiffusionDet}, {YOLOv8} and {YOLOv5} achieve moderately higher detection accuracy than their counterparts. In contrast, the proposed {RSTNet} model successfully detects and correctly labels all five nearshore ships, with neither misses nor false positives. These results indicate that {RSTNet} possesses superior nearshore detection capability compared with the other nine models, highlighting its strong suitability for SAR based ship detection on the {SSDD} dataset.

\begin{figure*}[t]
    \centering
    \includegraphics[width=0.8\linewidth]{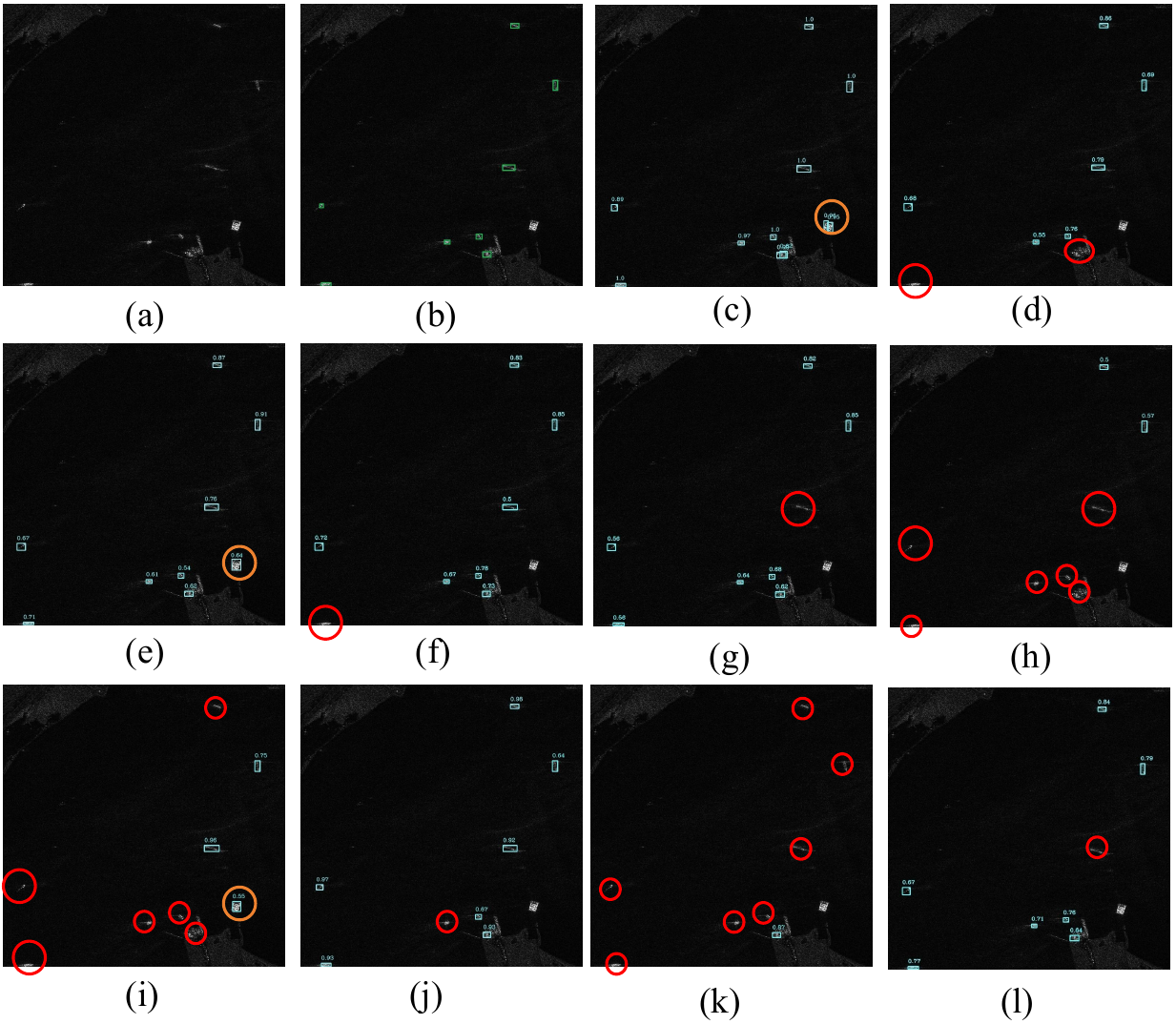}
    \caption{Revised comparison chart: Performance of RSTNet and other models on the HRSID dataset. (a) Original image. (b) Ground Truth. (c) C-AFBiFPN visualization results. (d) YOLOv10 visualization results. (e) DiffusionDet visualization results. (f) YOLOv8 visualization results. (g) YOLOv5 visualization results. (h) FCOS visualization results. (i) RetaiNet visualization results. (j) SSD visualization results. (k) Faster Rcnn visualization results. (l) RSTNet visualization results. The Green rectangles indicate GT, blue rectangles indicate predicted results, red ellipses indicate missed detections and orange circles indicate false detections.}
    \label{HRSID result}
\end{figure*}

Fig.~\ref{HRSID result} presents a comparative visualization of the detection performance for {RSTNet} and nine other models on the {HRSID} dataset. Fig.~\ref{HRSID result} (a) shows the original SAR image, Fig.~\ref{HRSID result} (b) displays the ground truth annotation with eight ship targets (three inshore small ships and five offshore small ships), and Fig.~\ref{HRSID result} (c)–(l) illustrate the detection results of {C-AFBiFPN}, {YOLOv10}, {DiffusionDet}, {YOLOv8}, {YOLOv5}, {FCOS}, {RetinaNet}, {SSD}, {Faste R-CNN} and {RSTNet}. Visual analysis reveals that {C-AFBiFPN} and {DiffusionDet} correctly detect nine ships (marked with blue rectangles), but they incur one false positive (FP, marked with orange circles) and no miss detections (marked with yellow triangles). {RetinaNet} achieve only two correct detections, accompany by one FP and six missed ships. Three models: {FCOS}, {RetinaNet} and {Faster R-CNN} exhibit the lowest detection accuracy, as they consistently fail to identify inshore ships and frequently misclassified offshore ships as background. {YOLOv10} correctly detects six ships but misses two; {YOLOv8}, {DiffusionDet} and {SSD} each correctly detects seven ships with one miss. Collectively, these three models achieve moderately higher accuracy than the aforementioned lower performing ones. In contrast, the proposed {RSTNet} model successfully detects and correctly labels seven of the eight target ships, with only one failure. These results demonstrate that {RSTNet} possesses superior capability in detecting inshore and offshore small ships compared to the other nine models, highlighting its strong suitability for SAR based ship detection on the {HRSID} dataset.

In summary, the performance of {RSTNet} is noteworthy, as it consistently outperforms other models on {SSDD} and {HRSID} datasets under complex environmental conditions, thereby demonstrating its effectiveness for SAR ship detection.

\section{Conclusion}
In this paper, we propose {RSTNet}, a detection framework tailored for SAR ship detection in challenging conditions, including non-uniform illumination, heavy noise, and small targets. We evaluate {RSTNet} on two public benchmarks, {HRSID} and {SSDD}, and compare it against nine representative detectors, including {C-AFBiFPN}, {YOLOv10}, {DiffusionDet}, {YOLOv8}, {YOLOv5}, {FCOS}, {RetinaNet}, {SSD} and {Faster R-CNN}. Experimental results show that {RSTNet} achieves {88.9\%} mAP on {HRSID} and {97.3\%} mAP on {SSDD}, outperforming competing methods in overall detection accuracy. In addition, {RSTNet} yields consistent gains on mAP@0.5:0.95, indicating improved robustness under stricter localization criteria and varying IoU thresholds. These results demonstrate that {RSTNet} effectively handles complex SAR backgrounds and improves detection quality for small ships, making it suitable for high-precision maritime monitoring applications. 
Despite its strong performance, {RSTNet} still incurs relatively high computational overhead and shows reduced practicality in resource-constrained settings. Future work will therefore explore more lightweight architectural designs and more efficient training and inference strategies, including anchor-free formulations, to further improve efficiency while preserving detection accuracy.

\section*{Author Contributions}
Conceptualization, X.Z., S.L., Z.C. and Y.Z.; resources, Z.C.; methodology, J.C.; validation, T.Y.; formal analysis, H.N.; investigation, H.N.; data curation, J.C. and P.H.; writing—original draft preparation, X.Z., S.L., Y.Z. and J.C.; writing—review and editing, X.Z., S.L., Y.Z.; visualization, P.H. and H.N.; supervision, H.N. and Z.C.; funding acquisition, X.Z., S.L. and Y.Z. All authors have read and agreed to the published version of the manuscript.
\section*{Funding}
The work was supported by Key Discipline Project on Electronic Information in Henan Province; the Low-altitude Agricultural Remote Sensing Team Project of Anyang Institute of Technology; Anyang Institute of Technology Doctoral Research Start up Fund Project: Research on Explainable Artificial Intelligence Methods Based on Big Data and Knowledge Graph (BSJ2023012); Research Project of the Ministry of Education on Experimental Teaching and the Construction of Teaching Laboratories: Reform and Practice of the Experimental Practice Teaching System of Electronic Information Based the ”Three-Four-Three” Model (SYJX2024-022); Ministry of Education Curriculum and Teaching Materials Research Project (2025): Comparative Study on Domestic and International Engineering Education Curriculum and Textbook Development (25BC0027); Tianjin Continuing Education Teaching Reform and Quality Improvement Research Program (2025): Research on Innovation and Development of University Continuing Education and Lifelong Learning Driven by Information Literacy and Technological Foresight Capability (J2025001); Tianjin Higher Education Postgraduate Education Reform Research Program: Research and Practice on a Multi-Dimensional Education Model for Electronic Information Professional Degree Postgraduates Based on "Dual Collaboration {\&} Four Integrations" (TJYGZ44). 
\section*{Data Availability Statement}
The data presented in this study are available on request from the corresponding author, as the mine data are subject to certain confidentiality requirements.
\section*{Conflicts of Interest}
The authors declare no conflicts of interest.

\bibliography{sample}
\end{document}